\documentclass[conference]{IEEEtran}
\IEEEoverridecommandlockouts

\usepackage{cite}
\usepackage{amsmath,amssymb,amsfonts}
\usepackage{algorithmic}
\usepackage{graphicx}
\usepackage{textcomp}
\usepackage{xcolor}
\usepackage{subcaption}
\usepackage{url}

\usepackage{times}
\usepackage[labelformat=simple]{subcaption}

\usepackage{multirow}
\usepackage{float}
\usepackage{wrapfig}
\usepackage[most]{tcolorbox}

\def\BibTeX{{\rm B\kern-.05em{\sc i\kern-.025em b}\kern-.08em
    T\kern-.1667em\lower.7ex\hbox{E}\kern-.125emX}}
\begin{document}

\title{Automatic Tag Recommendation for\\Painting Artworks Using Diachronic Descriptions
\thanks{We thank the partial support given by the Project: Models, Algorithms and Systems for the Web (grant FAPEMIG / PRONEX / MASWeb APQ-01400-14), and authors' individual grants and scholarships from CNPq, Fapemig and Kunumi.}
}

\author{\IEEEauthorblockN{Gianluca Zuin}
\IEEEauthorblockA{
\textit{CS Dept. UFMG}\\
Belo Horizonte, Brazil\\
gzuin@dcc.ufmg.br}
\and
\IEEEauthorblockN{Adriano Veloso}
\IEEEauthorblockA{
\textit{CS Dept. UFMG}\\
Belo Horizonte, Brazil\\
adrianov@dcc.ufmg.br}
\and
\IEEEauthorblockN{Jo\~ao C\^andido Portinari}
\IEEEauthorblockA{
\textit{Projeto Portinari}\\
Rio de Janeiro, Brazil\\
portinari@portinari.org.br}
\and
\IEEEauthorblockN{Nivio Ziviani}
\IEEEauthorblockA{
\textit{CS Dept. UFMG \& Kunumi}\\
Belo Horizonte, Brazil\\
nivio@dcc.ufmg.br}
}

\maketitle

\begin{abstract}
In this paper, we deal with the problem of automatic tag recommendation for painting artworks. Diachronic descriptions containing deviations on the vocabulary used to describe each painting usually occur when the work is done by many experts over time. The objective of this work is to provide a framework that produces a more accurate and homogeneous set of tags for each painting in a large collection. To validate our method we build a model based on a weakly-supervised neural network for over $5{,}300$ paintings with hand-labeled descriptions made by experts for the paintings of the Brazilian painter Candido Portinari. This work takes place with the Portinari Project which started in 1979 intending to recover and catalog the paintings of the Brazilian painter. The Portinari paintings at that time were in private collections and museums spread around the world and thus inaccessible to the public. The descriptions of each painting were made by a large number of collaborators over 40 years as the paintings were recovered and these diachronic descriptions caused deviations on the vocabulary used to describe each painting. Our proposed framework consists of (i) a neural network that receives as input the image of each painting and uses frequent itemsets as possible tags, and (ii) a clustering step in which we group related tags based on the output of the pre-trained classifiers.
\end{abstract}


\section{Introduction}

Image and tag annotation is a problem that has been thoroughly studied in the literature \cite{imageanotation,ref12}. Given an image, we wish to associate a set of tags that best describes and summarizes the image. Traditional methods focus on using human-labeled images as training data and obtaining models that learn relationships between the image and the closed set of concise candidate tags~\cite{ref11}. However, a significant amount of real-world problems 
contain unlabeled, weakly labeled or raw text data. To tackle the first task, we need first to mine the available resource in the search for suitable annotations which might be not trivial. One particular case is when we do not have a concise set of annotations that summarize an image but rather, a rich text that thoroughly describes each of its aspects.

\begin{figure}[t]
\centering
\includegraphics [width = 1\linewidth] {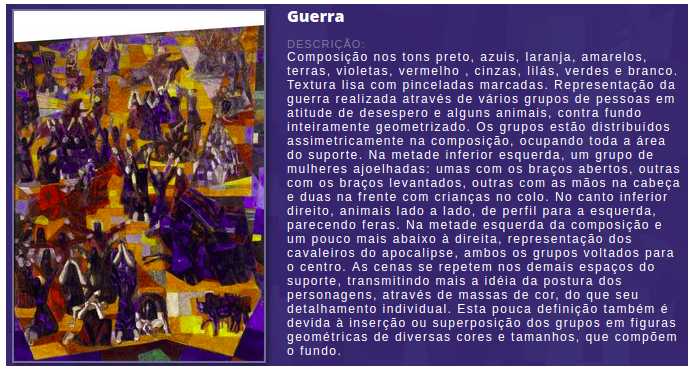}

\caption{Painting entitled ``War'', and its description taken from the Portinari Project website. It measures 14.32 meters (47.0 ft) tall and 10.66 meters (35.0 ft) large. ``War and Peace'' (Portuguese: ``Guerra e Paz'') are two paintings by Candido Portinari painted between 1952 and 1956 for the United Nations Headquarters as a gift from the Brazilian government.}

\label{fig:pinturadesc}
\end{figure}

This work takes place in partnership with the \emph{Portinari Project} and has as its main goal testing our solution for the two aforementioned tasks. The dataset employed consists of $5{,}300$ Portinari paintings with descriptions made by artwork experts. As an example, Figure \ref{fig:pinturadesc} illustrates a painting and its associated description.

The process of describing Portinari artworks was performed by many collaborators over a large period of time, leading to diachronic data and divergent tags. For example, the set of words ``nose, eyes, eyebrow, chin'' and `` mouth, cheeks, pupils'' both represent a face but may exclusively appear in different classifications. Deep neural networks can learn the concepts associated with a face, for instance, and should recommend tags within both of these sets~\cite{ref2}.

The Portinari Project\footnote{\url{www.portinari.org.br}, \url{http://artsandculture.google.com/project/portinari}} was created in 1979 and since then it has recovered and cataloged over $5{,}300$ paintings, drawings, and engravings, as well as more than 30 thousand documents regarding the work of the Brazilian painter Candido Portinari (1903, 1962).
The project also aims to provide a view of Brazilian cultural identity and preserving its national memory while also spreading the humanistic and ethical message conveyed through Portinari's paintings, writings, and poems to the broader population \cite{lanzelotte1993portinari}.

We propose to solve the tasks of tag creation and recommendation by mixing frequent itemset mining methods \cite{Han00miningfrequent} with state-of-the-art deep neural networks trained upon a convolutional autoencoder \cite{poultney2007efficient,ref6,david2016deeppainter}. We tackle the task of tag refinement over the set of candidate tags to select a concise and highly informative set. By extracting itemsets we can get an overall context of a tag given co-occurring words~\cite{ref7}. It should be highlighted that one of Portinari Project main goals is the spread of Portinari's message conveyed through his art. The knowledge found in itemsets may be vastly more helpful to experts and the broader audience alike when observing a painting~\cite{ref1}. Regarding autoencoders, their usage has been explored extensively in the literature and provides a useful method for feature extraction and data compression with minimum loss of information~\cite{ref9}. Finally, we propose grouping together similar itemsets according to the classifiers' output to obtain a concise and highly informative set of tags.

Although we focus upon the particular problems found in the Portinari dataset, the techniques described are general enough that they should be able to be applied to any other problem that shares similar characteristics. Namely, divergent tags and the presence of textual data to extract image labels. This work proposes not only a tool to classify paintings but also a mean to generate tags from raw descriptions. In practice, we claim the following contributions:
\begin{itemize}
\item We propose a deep neural network trained over the multi-label classification, allowing us to use descriptions from previous paintings to recommend tags for new ones.
\item We propose a novel method for regularization and refinement of tags.
\item We expand the original dataset by creating compact representations, aiding further research on the subject.
\item We expand the research performed by the \emph{Portinari Project} in the hope that the knowledge acquired will help in its socio-cultural and educational mission.
\end{itemize}

\section{Related Work}

The image-tag annotation task can be defined as the process in which a computer automatically specifies a tag to a digital image. Classical approaches involve the usage of SVM, KNN or both~\cite{7175695}. However, convolutional neural networks have been proven to provide superior results \cite{ojha2017image}. For instance, Krizhevsky et al. \cite{KrizhevskySH17} reported state-of-the-art results on ILSVRC 2012 which contains over 1,000 categories. Zhang et. al \cite{twobirds2018} address not only the task of image annotation but also tag-refinement through the usage of a deep neural network. The proposed approach is based in the following three premises: (a) users provide a set of tags biased towards their perception; (b) images are marked with a small number of tags; (c) images with similar visual appearance are usually annotated with the same tags. By manipulation of training batches and image semantic neighborhoods, the authors trained a model to predict correlated tags for similar images through weakly supervised learning. The main difference from his work and ours lies in the tags themselves. Since we only have the descriptions of the paintings to work upon, we need first to create our tags. This leads to paintings having wide ranges of labels and some tags describing more specific and in-depth aspects in some paintings than in others, in disagreement with premises (b) and (c). Also, since the descriptions were made by artwork experts following a set of rules, the bias mentioned in (a) is mitigated and mostly arises from stylistic preferences.

The idea of using autoencoders to learn relevant features was originally proposed in \cite{poultney2007efficient}. The authors considered the problem of learning complete representations of the input data. That is representations that preserve the information contained in the input and are capable of reconstructing it. They established a framework based on learning to compress data from the input layer into a short code, and then uncompress that code into something that closely matches the original data. The output of the coding section, which consists of a compressed representation of the input, contains enough information to summarize and reconstruct the original data. Their work provides the theoretical framework and the mathematical basis that justify this kind of approach. Since then, similar approaches have emerged in which autoencoders are used as a means of learning relevant features regarding the input data. Gao et. al. \cite{7124463} used autoencoders to learn image representation which is then used as input for a face recognition system. Krizhevsky and Hinton \cite{dblp1829095} take advantage of the compact intermediary output of autoencoders to map images to binary codes, which are then used for image retrieval.

Among works consisting of autoencoders applied to paintings, we mention the one by David and Netanyahu \cite{david2016deeppainter} which tackles the task of painter classification. They divide the problem into two steps. The first one is the training of a convolutional autoencoder on a dataset consisting of 5{,}000 paintings extracted from the Web museum. The second is training a classifier over the intermediary output of the autoencoder, identifying if a painting belongs to Rembrandt, Renoir or van Gough. The classifier proposed to consist of the encoding step of the autoencoder followed by fully connected layers and achieves an accuracy score of 96.52\%. This model is similar to our employed neural network.

In contrast, we are interested in solving a multi-class multi-label classification problem. Regarding this subject, Pu et. al. \cite{NIPS2016_6528} train a convolutional autoencoder upon a set of images. Its latent representation is then mapped to a set of labels and captions using bayesian SVMs and recurrent networks. They reported results that are comparable to the state-of-the-art on the evaluated datasets. Our work goes one step further by finding frequent itemsets to be used as possible tags for a given painting. Also, we model our classifier as a multi-layer perceptron network instead of a bayesian SVM to allow the gradient descent step to update the weights of the encoding step. The main goal of the employed autoencoder is to learn through pre-training relevant features. By allowing its weights to be updated, we fine-tune it to the task at hand. We include SVM and KNN solutions as a baseline in our experiments, showing the benefits of our approach.

\begin{figure*}
\centering
\includegraphics[width=0.7\linewidth]{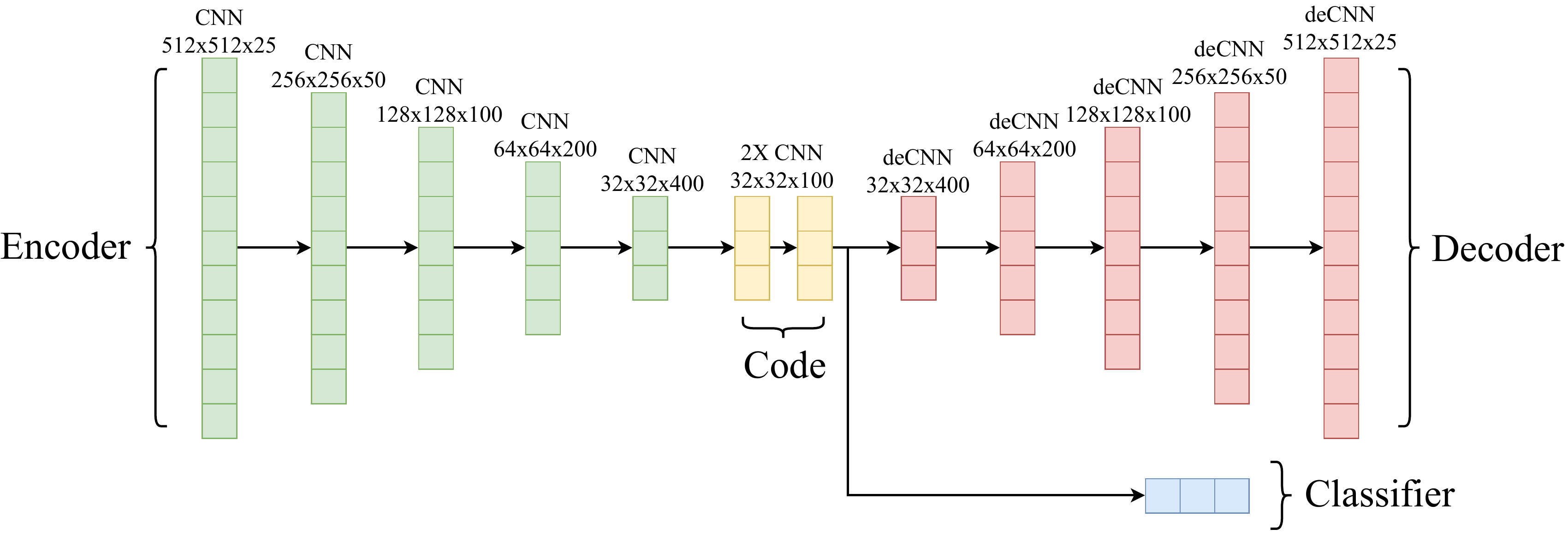}
\caption{Proposed neural network for the classification task. The encoder output is used as input for the classifier.}
\label{fig:class}
\end{figure*}

\section{Proposed approach}
\label{sec:classcnn}

We formulate our classification model as a function $f(p,T;\theta)$ parameterized by $\theta$ that maps a painting to a set of scores associated with each candidate tag. The candidate tags (aka, classes) are extracted though itemset mining, and then paitings are classified into these classes~\cite{hwang2012reading}. Formally, given a painting $p$ and a list of candidate tags $T = {t_1,t_2,...,t_n}$, we assume two  learning scenarios: one where we learn a single $f(p,T;\theta)$ and a second in which we learn multiple $f_i(p,t_i;\theta)$ to calculate the relevance between $p$ and every candidate $t_i \in T$.

The first scenario employs a one-to-many classification. Given the model's output, we perform tag refinement by grouping tags. This approach is reliant on the interdependency of features. Similar tags should focus on similar image characteristics and, consequently, share a prediction over the same paintings regardless of the ground truth annotation. This leads to a concise set of tags that attempt to solve the problem of divergent diachronic tags. The second scenario consists of a one-to-one classifier trained on each class individually. Since the models were trained independently, we cannot take advantage of tag refinement by exploiting the interdependency of features. Rather, we evaluate a tag's suitability as a candidate label by assessing its frequency and the validation loss found after cross-validation. This enables us to filter out labels that do not characterize well the data and leads to a set of highly concise, informative and discrete classes.

\vspace{0.1in}
\noindent \textbf{Itemset mining:} Itemset mining allows us to verify which and how often a given set of objects of interest co-occur. Let $I = {i_1, i_2, \ldots , i_m}$ be a set of elements called items. For instance, the set of all the words contained in the descriptions of Portinari's paintings. An itemset is a set in which all its elements are contained in $I$. Further, let $D = {d_1, d_2, \ldots , d_n}$ be a set of elements called transaction identifiers, that is, an identifier to every description contained in the dataset. We denote the support of an itemset $X$ as the number of transactions in which it occurs. The relative support of $X$ is an estimate of the joint probability of the items contained in $X$ and can be found by the ratio of transactions with the occurrence of $X$. The problem of frequent itemset mining consists in the retrieval of all itemsets that have relative support not lower than a given threshold, which can be solved by employing the FPGrowth algorithm \cite{Han00miningfrequent}, where we build a prefix tree in which child nodes correspond to extensions of the root item and store its respective support value.

\vspace{0.1in}
\noindent \textbf{Feature learning:} The proposed architecture consists of a neural network that receives as input the paintings and returns a vector of probabilities. Each of the frequent itemsets found in the description of the paintings represents a position in this vector. We want tags that appear in the painting to have high probability values, while those that do not appear to assume low values. The features used in our classifier are learned with a convolutional autoencoder over the paintings contained in the dataset. The latent representation derived from its intermediary output consists of a summarization of the painting and is fed to a multi-layer perceptron network. The last layer gives us the probabilities of each tag being present in the input image. Figure \ref{fig:class} illustrates the proposed network.

\begin{figure}
\centering
\includegraphics[width=\linewidth]{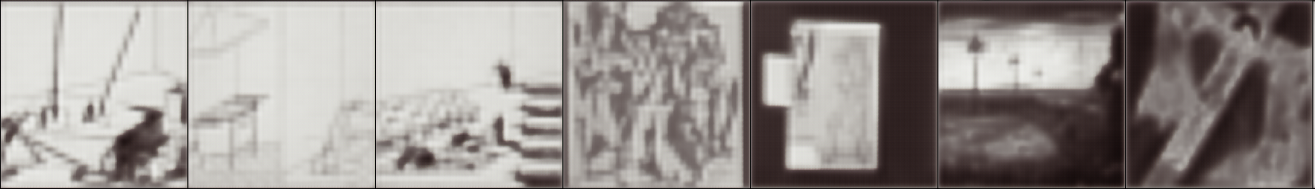}
\includegraphics[width=\linewidth]{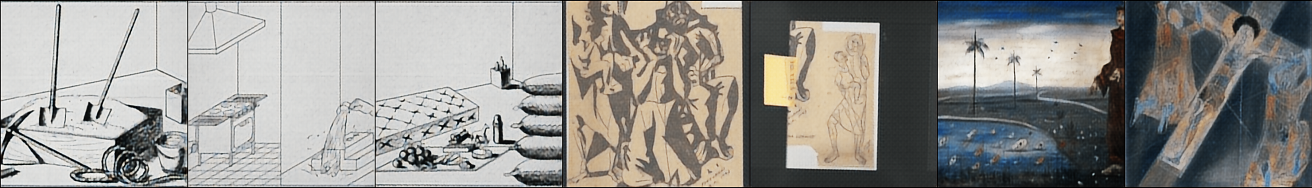}
\includegraphics[width=\linewidth]{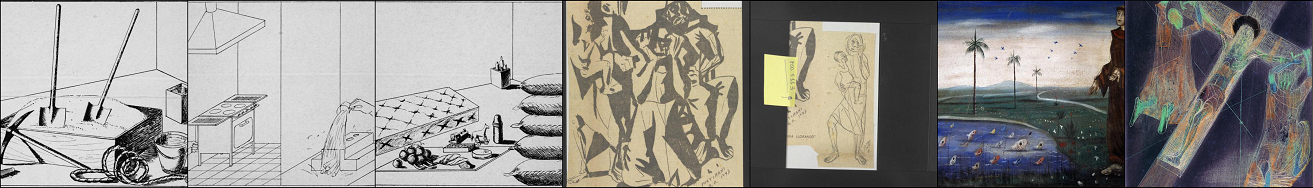}
\caption{Output of the convolutional autoencoder after training for one epoch, 100 epochs and the original artworks.}
\label{fig:autoencoderout}
\end{figure}

 The core of our architecture lies in the autoencoder and its ability to create rich and compact representations. Figure \ref{fig:autoencoderout} shows the output of the convolutional autoencoder and the original painting. At the start of training, the model is only able to create blurs that loosely resemble the original paintings. After 100 epochs, it learns to recreate the input images, although there is a small loss concerning vibrant colors.

\vspace{0.1in}
\noindent \textbf{Meta-tag acquisition:} In order to group similar tags, we employ the standard K-means algorithm \cite{forgy65}. Formally, given a set of observations $(x_1, x_2, ..., x_m)$, where each observation represents a $d$-dimensional vector, the clustering method attempts to find a partition of these $m$ observations into $k$ groups $\{G_1, G_2, \ldots, G_k\}$ that minimize an instability metric within these groups. In our work, we interpret each group as a meta-tag summarizing all observations (tags) contained within. Determining the number of clusters in a dataset is a distinct issue from the process of actually solving the clustering problem. One possible method for selecting $k$ is through a cluster's silhouette \cite{rousseeuw1987silhouettes}. Silhouette analysis provides an interpretation of consistency within clusters of data by measuring the separation distance between the generated clusters.

\begin{figure}[t]
\centering
\includegraphics[width=.7\linewidth]{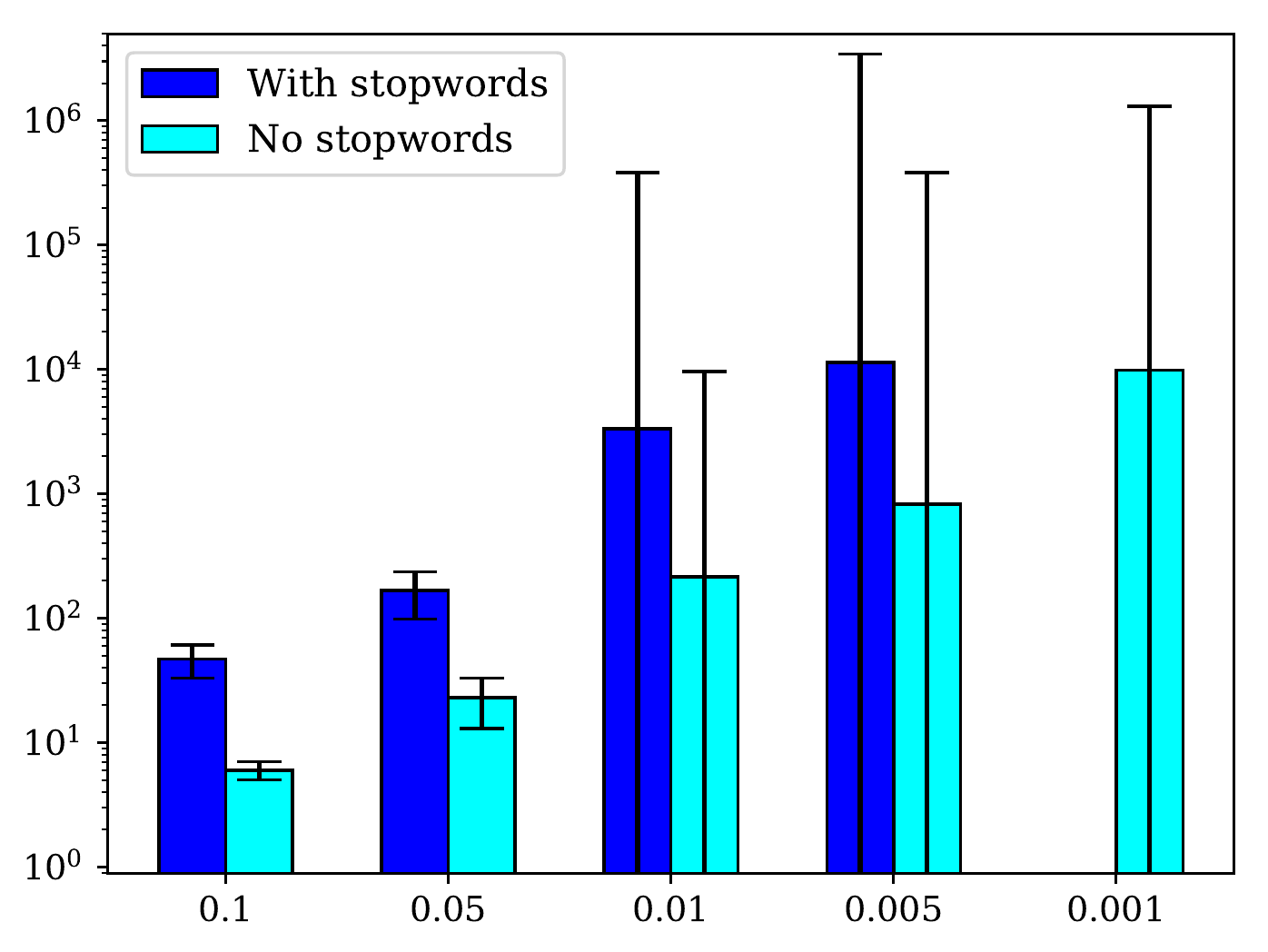}
\caption{Average number of patterns found in each painting's description. Chart in logarithmic scale.}
\label{fig:patterns}
\end{figure}

\section{Experiments}
\label{sec:exp}

Different levels of minimum relative support within the range $ [0.1, 0.001]$ were investigated. Processing frequent itemsets with low values of minimum support became unfeasible due to the large combination of possible tags.
The following research questions aim to be answered throughout our experiments:

\begin{description}
\item[RQ1:] What is the impact of stopwords during frequent itemset mining?
\item[RQ2:] Is frequent itemset mining effective for creating meaningful tags?
\item[RQ3:] Which are relevant values of minimum relative support?
\item[RQ4:] Is it possible to get a more concise set of tags using the output of a pre-trained classifier?
\end {description}

\subsection{Data}
Figure \ref{fig:patterns} shows the number of itemsets found in the descriptions by maintaining and removing stopwords. It is possible to observe that there are fewer sets in the models in which we remove stopwords as expected. We also compute the mean number of tags per painting and their confidence interval. We treat each sentence as a transaction. In this experiment, we wish to visualize the average presence of frequent itemsets throughout the set of sentences that form a description. Smaller values of minimum relative support lead to high variance, possibly indicating that the classifier will show poor performance.

\begin{figure}[t]
\centering
\includegraphics[width=.7\linewidth]{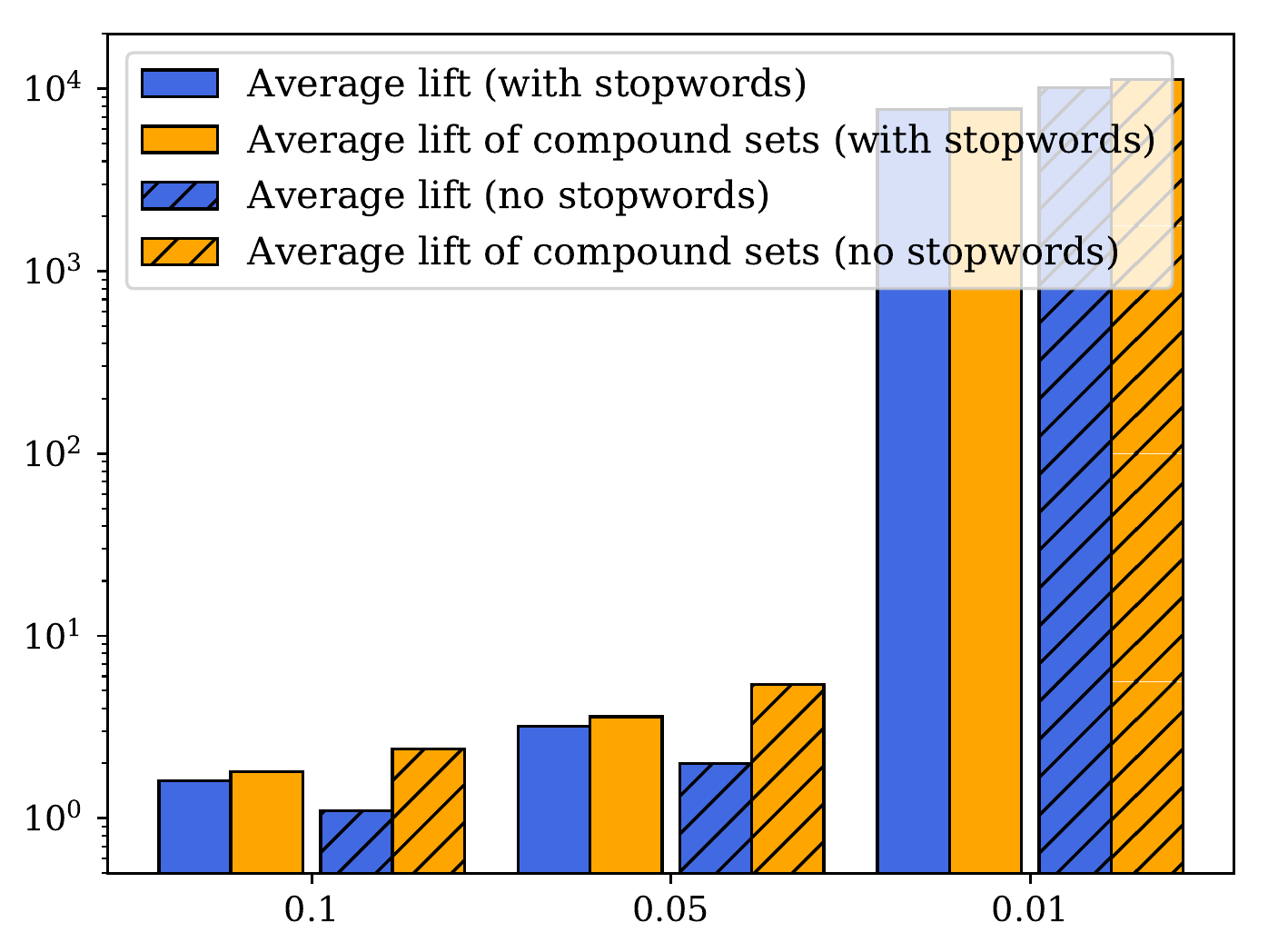}
\caption {Average lift of frequent itemsets while keeping and removing stopwords. Chart in logarithimic scale.}
\label{fig:liftpadroes}
\end{figure}

Regarding RQ2, Figure \ref{fig:liftpadroes} shows the mean lift of the frequent itemsets in comparison to the compound itemsets. According to Zaki and Meira Jr. \cite{zaki}, the lift of a rule is defined as the ratio between observed joint probability and the expected joint probabilities assuming that the items are independent. A value close to 1 means that the pattern is expected considering the support of its components. Since the frequent compound sets have higher lift values than the overall frequent itemsets, we may conclude that they increase the lift value of the frequent itemsets. This corroborates with the insight that frequent itemsets provide tags with more information than simply using the most frequent words.

\begin{table}[h]
\centering
\begin{tabular}{lc|rrr}
\hline
\multicolumn{1}{c}{} & Min sup. & P & R & $F_1$ \\ \hline
\multirow{3}{*}{\textit{No stopwords}} & 0.01 & .620 & .102 & .175 \\
& 0.05 & .599 & .527 & .561 \\
& 0.10 & .762 & .855 & .809 \\ \hline
\multirow{3}{*}{\textit{With Stopwords}} & 0.01 & .568 & .064 & .115 \\
& 0.05 & .770 & .705 & .736 \\
& 0.10 & .798 & .857 & .826 \\ \hline\\
\end{tabular}
\caption{Performance of the Autoencoder-MLP* classifier.}
\label{tab:classifier}
\end{table}

\begin{figure*}[t]
\centering
\includegraphics[width=0.25\linewidth]{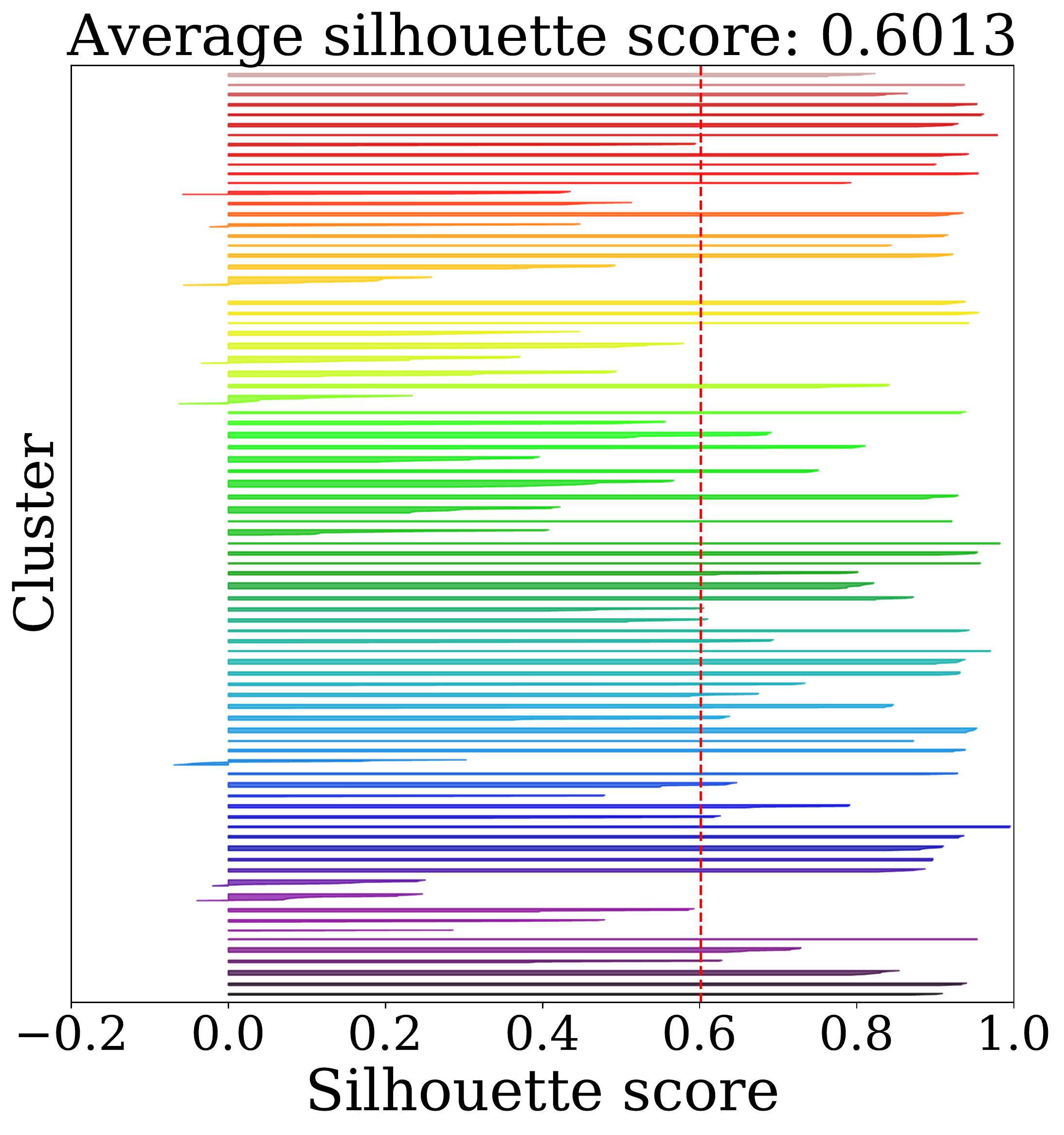}
\includegraphics[width=0.25\linewidth]{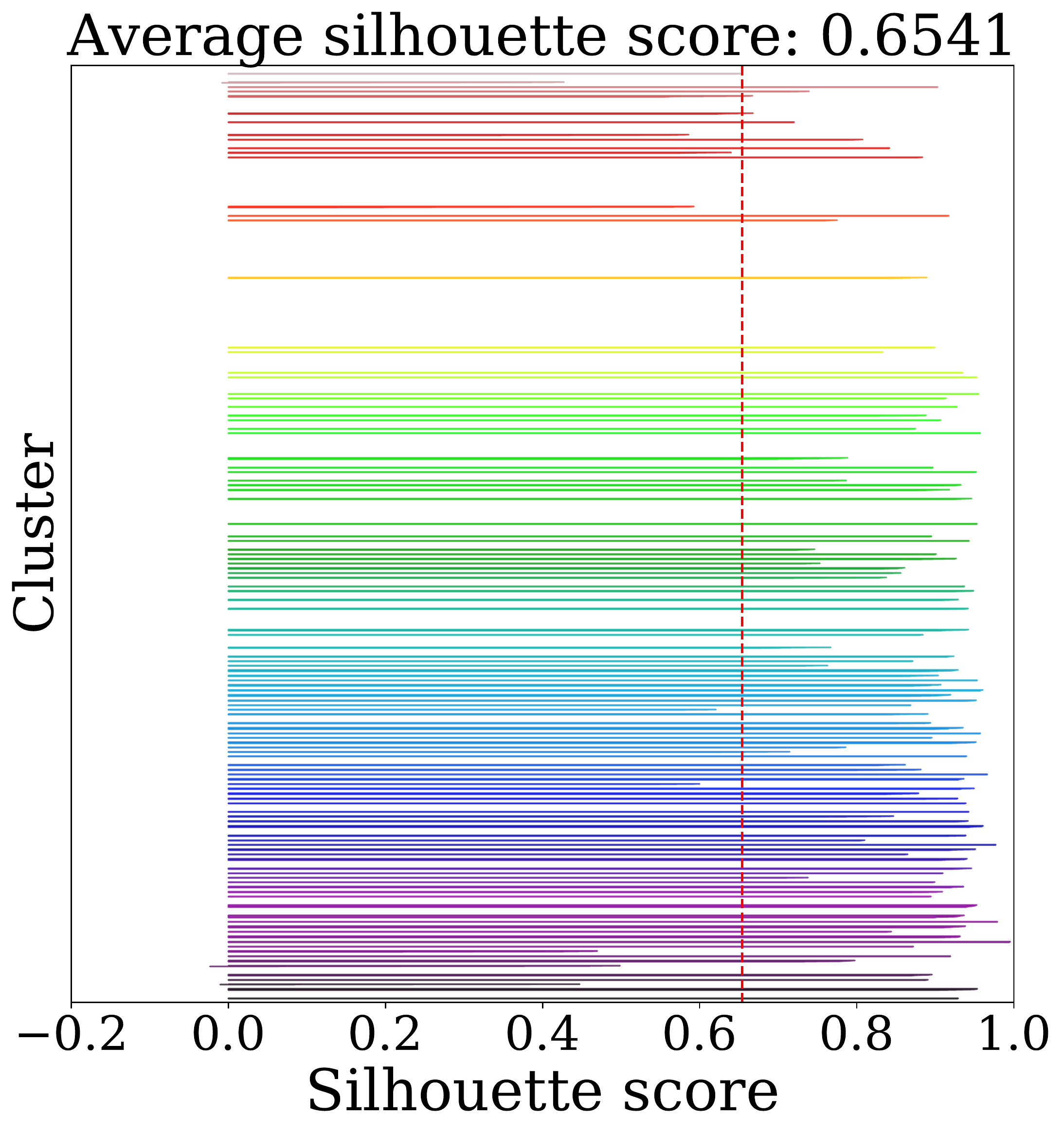}
\includegraphics[width=0.25\linewidth]{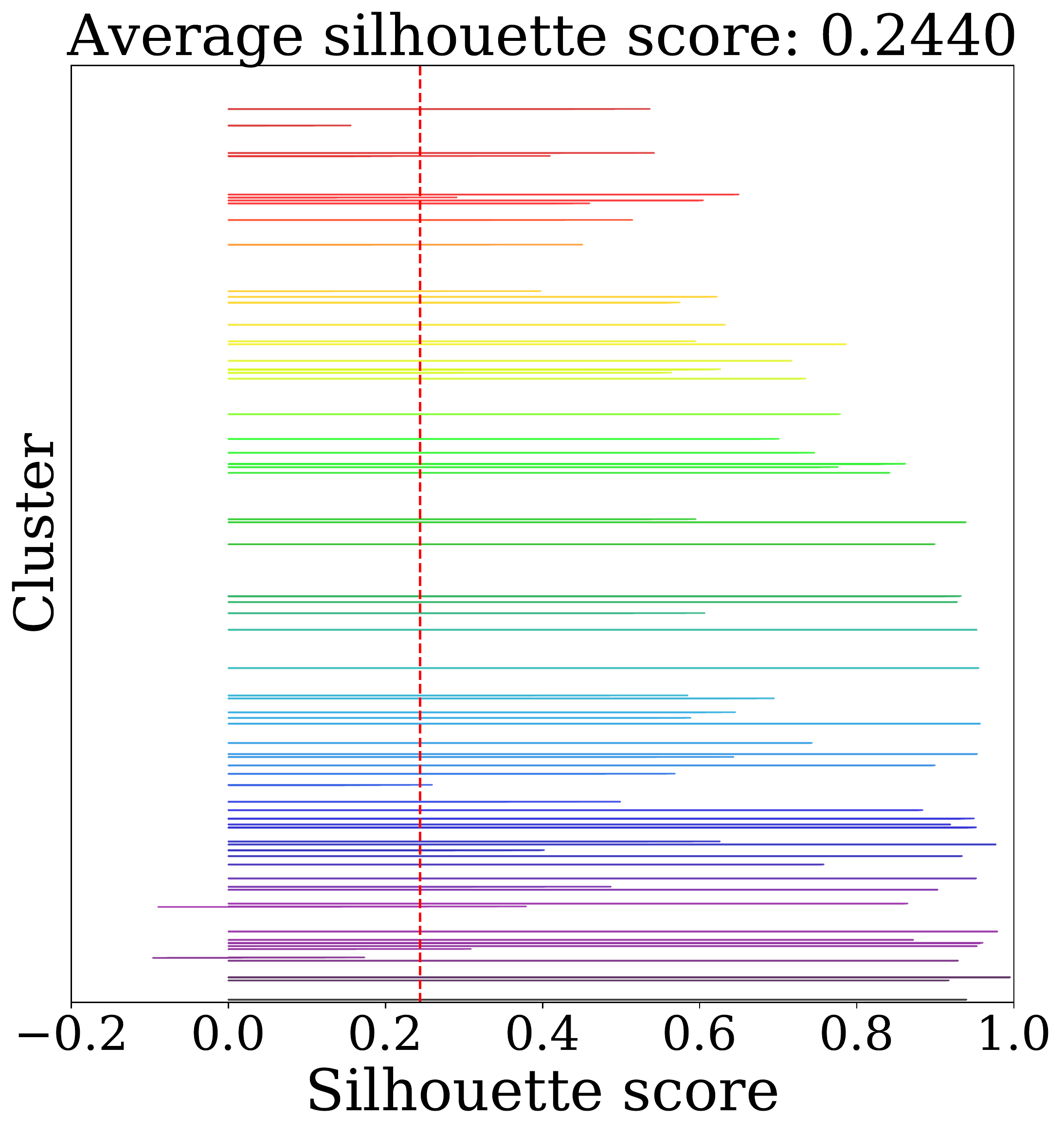}
\caption{Silhouette analysis for very small and very large $k$ (increasing values of $k$ go from left to right). Each color represents a different cluster while the dashed line gives the average Silhouette score.}
\label{fig:silluete}
\end{figure*}

\subsection{One-to-many classifiers}
\label{sec:onemany}

We previously observed that high support leads to few itemsets, and low supports are associated with high variance in the number of tags per painting. We can answer RQ3 by considering the two scenarios in which the autoencoder latent representation is used as input for the classification, one where the weights of the encoding step are updated {(Autoencoder-MLP*)} and one where only the weights from the classification step are trained (Autoencoder-MLP). Table \ref{tab:classifier} shows the micro-precision and micro-recall values obtained with Autoencoder-MLP* for each minimum relative support threshold. 

We also explore if using an autoencoder to reduce the input size is preferable to more direct approaches such as scaling the input image. We evaluate the performance of classifiers in which we downscaled the images to match the autoencoder compression rate and also employ other classical methods for image classification found in the literature. For a random guess baseline, we randomly give a prediction matching the occurrence probability of each given tag in the 'training' split. For further validation, the random guess baseline is repeated 30 times and the average result is presented. 

\begin{table}[h]
\centering
\begin{tabular}{lrrrrrr}
\hline
& \multicolumn{3}{r}{With stopwords} & \multicolumn{3}{r}{No stopwords} \\
Minimum Support & 0.10 & 0.05 & \multicolumn{1}{c|}{0.01} & 0.10 & 0.05 & 0.01 \\ \hline\\[0.05ex]
KNN & .808 & .633 & \multicolumn{1}{c|}{.270} & .784 & .527 & .150 \\
SVM & .786 & .642 & \multicolumn{1}{c|}{.292} & .774 & .546 & .200 \\
MLP & .815 & .683 & \multicolumn{1}{c|}{.241} & .803 & .555 & .183 \\
Autoencoder-KNN & .851 & .678 & \multicolumn{1}{c|}{.279} & .802 & .554 & .205 \\
Autoencoder-SVM & .832 & .681 & \multicolumn{1}{c|}{.299} & .807 & .563 & .262 \\
Autoencoder-MLP & .898 & .738 & \multicolumn{1}{c|}{.243} & .832 & .560 & .242 \\
Autoencoder-MLP* & .826 & .736 & \multicolumn{1}{c|}{.115} & .809 & .561 & .175 \\ \hline
Random Guess & .602 & .524 & \multicolumn{1}{c|}{.104} & .605 & .459 & .132 \\ \hline \\
\end{tabular}
\caption{$F_1$-$score$ values of evaluated classifiers. The usage of a autoencoder representation shows significant improvements across the board.}
\label{tab:knn}
\end{table}

Given the nature of the data, models often return a superset or subset of the ground-truth labels. There are also many instances in which the model returns predictions with semantics similar to the expected outputs. Thus, we can infer that models learn general painting features, which leads to tags with similar meanings being predicted in similar paintings. We propose obtaining clusters related to each semantic sense and use them as meta-tags for more accurate classification. Given the painting set $P$ and the candidate tag set $T$, we can get a representation of size $|P|$ for each of the $t \in T$ candidate tags consisting of the predictions returned by the classifier. Again, we employ the standard K-Means algorithm to cluster these representations and perform silhouette analysis as an estimator of the best value of $k$. We explore large numbers of clusters in comparison to literature to find concise yet sizeable groups of meta-tags. The evaluated $k$ ranges from 10\% up to 90\% of the total data size. Figure \ref{fig:silluete} shows the silhouette values of some explored scenarios. For very large $k$, we have an abundance of clusters with only a single tag which leads to a silhouette of 0 as shown in the rightmost chart.

We retrain the classifier after retagging the paintings according to the new meta-tags derived from the clusters. Tables \ref{tab:mlp-005} to \ref{tab:mlp001_2} summarize the performance of the classifier in terms of Precision (P), Recall (R) and $F_1$-$score$ as we reduce the tag-space size. $RatioF_1$ states how much better the classifier is in comparison to the random guessing baseline. There is no statistical difference between the original model and the classifiers trained with only a small reduction in space. Large reductions, however, lead to larger gains in performance.

\begin{table}[!htb]
\centering
\begin{tabular}{lrrrrr}
\hline
Tag-space & P & R & $F_1$ & $RatioF_1$ & Silhouette \\ \hline\\[0.05ex]
\textit{100\%} & \textit{.770} & \textit{.705} & \textit{.736} & \textit{1.402} & \textit{-} \\
90\%	&.780	&.655	&.712	&1.389	&.129 \\
80\%	&.794	&.631	&.703	&1.376	&.244 \\
70\%	&.798	&.621	&.698	&1.379	&.368 \\
60\%	&.801	&.616	&.696	&1.410	&.495 \\
50\%	&.777	&.645	&.705	&1.487	&.654 \\
40\%	&.765	&.691	&.726	&1.483	&.706 \\
30\%	&.797	&.679	&.733	&1.418	&.692 \\
20\%	&.819	&.814	&.816	&1.396	&.601 \\
10\%	&.881	&.885	&.883	&1.379	&.415 \\ \hline \\
\end{tabular}
\vspace*{-0.3cm}
\caption{Autoencoder-MLP* using stopwords and mininum support set to 0.05.}
\label{tab:mlp-005}
\vspace*{-0.1cm}
\end{table}

\begin{table}[!htb]
\centering
\begin{tabular}{lrrrrr}
\hline
Tag-space & P & R & $F_1$ & $RatioF_1$ & Silhouette \\ \hline\\[0.05ex]
\textit{100\%} & \textit{.740} & \textit{.737} & \textit{.738} & \textit{1.408} & \textit{-} \\
90\%	&.683	&.740	&.710	&1.415	&.059 \\
80\%	&.657	&.706	&.681	&1.416	&.090 \\
70\%	&.645	&.709	&.675	&1.417	&.119 \\
60\%	&.649	&.682	&.665	&1.415	&.150 \\
50\%	&.663	&.649	&.656	&1.401	&.175 \\
40\%	&.664	&.681	&.672	&1.406	&.199 \\
30\%	&.690	&.767	&.726	&1.411	&.226 \\
20\%	&.735	&.846	&.787	&1.404	&.234 \\
10\%	&.833	&.939	&.883	&1.377	&.212 \\ \hline \\
\end{tabular}
\vspace*{-0.3cm}
\caption{Autoencoder-MLP using stopwords and mininum support set to 0.05.}
\label{tab:mlp005_2}
\vspace*{-0.1cm}
\end{table}

\begin{table}[!htb]
\centering
\begin{tabular}{lrrrrr}
\hline
Tag-space & P & R & $F_1$ & $RatioF_1$ & Silhouette \\ \hline\\[0.05ex]
\textit{100\%} & \textit{.620} & \textit{.102} & \textit{.175} & \textit{1.348} & - \\
90\%	&.665	&.102	&.177	&1.330	&.084 \\
80\%	&.676	&.103	&.179	&1.334	&.190 \\
70\%	&.661	&.114	&.194	&1.419	&.323 \\
60\%	&.673	&.114	&.195	&1.403	&.442 \\
50\%	&.675	&.113	&.194	&1.417	&.493 \\
40\%	&.684	&.144	&.238	&1.460	&.507 \\
30\%	&.703	&.184	&.292	&1.527	&.513 \\
20\%	&.724	&.333	&.456	&1.664	&.590 \\
10\%	&.785	&.502	&.612	&1.702	&.604 \\ \hline \\
\end{tabular}
\vspace*{-0.3cm}
\caption{Autoencoder-MLP* without stopwords and minimum support set to 0.01.}
\label{tab:mlp-001}
\vspace*{-0.1cm}
\end{table}

\begin{table}[!htb]
\centering
\begin{tabular}{cccccc}
\hline
Tag-space & P & R & $F_1$ & $RatioF_1$ & Silhouette \\ \hline\\[0.05ex]
\textit{100\%} & \textit{.556} & \textit{.155} & \textit{.242} & \textit{1.847} & \textit{-} \\
90\%	&.581	&.146	&.233	&1.679	&.006 \\
80\%	&.547	&.170	&.259	&1.765	&.009 \\
70\%	&.499	&.219	&.304	&1.951	&.014 \\
60\%	&.586	&.179	&.274	&1.632	&.020 \\
50\%	&.589	&.203	&.302	&1.659	&.028 \\
40\%	&.612	&.222	&.326	&1.646	&.038 \\
30\%	&.600	&.270	&.372	&1.670	&.047 \\
20\%	&.614	&.370	&.462	&1.762	&.062 \\
10\%	&.654	&.605	&.629	&1.849	&.081 \\ \hline \\
\end{tabular}
\vspace*{-0.3cm}
\caption{Autoencoder-MLP without stopwords and minimum support set to 0.01.}
\label{tab:mlp001_2}
\vspace*{-0.1cm}
\end{table}

\subsection{One-to-one classifiers}

\begin{figure}[t]
\centering
\includegraphics[width=0.7\linewidth]{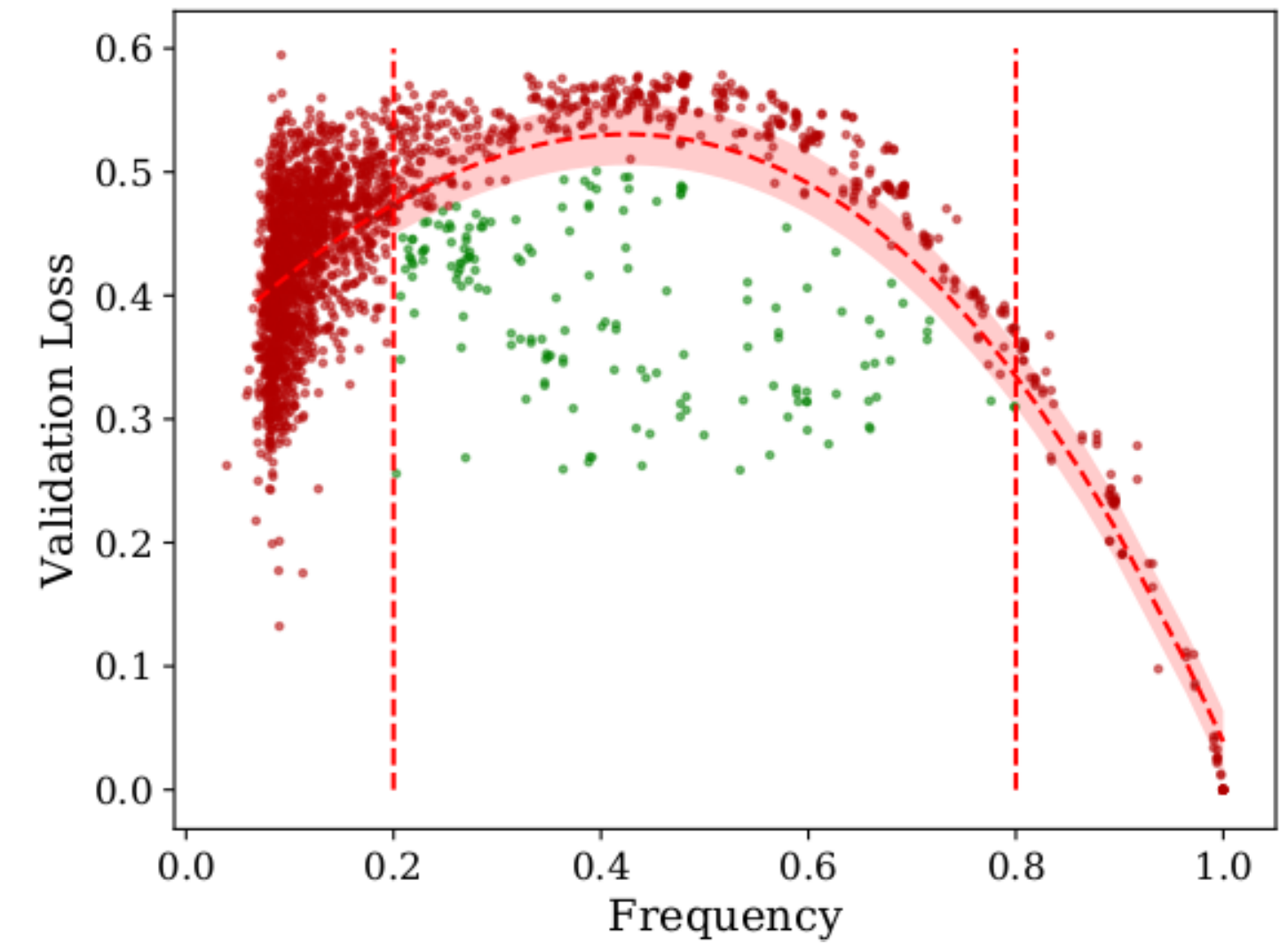}
\caption {Relationship between tag frequency and validation loss of a trained model. Tags below the red dashed curve consist of suitable tags for classification.}
\label{fig:valfreq}
\end{figure}

Our second approach consists of training a model to classify each candidate tag independently. One of the main drawbacks is that the model cannot use knowledge learned from other tags and exploit latent interdependency. However, we can perform an analysis in a case-by-case scenario and even filter out undesirable models. For instance, tags that are too frequent are not helpful as they do not discriminate well between paintings. Likewise, tags that are too infrequent are too rare to provide useful information. Overall, we wish to focus on tags that generalize well the data and provide high-performance models. Figure \ref{fig:valfreq} illustrates the distribution of candidate tags given their frequency and the validation loss found after 10-fold cross-validation of the model trained to predict them. The dashed lines represent the described constraints: not-frequent, not-rare and performant. If we only consider the red instances and plot a curve averaging each $\Delta$frequency, we should fit a curve that traverses the middle of the data points cloud and illustrates the average trend. By considering the green instances as well, this curve is naturally brought down and we can find a suitable cutting point between good and bad models. The decision boundary can be calculated by establishing a margin derived from the standard deviation.

We explore the effect of changing the minimum relative support threshold and the results are reported in Table \ref{tab:oneres}. It is important to highlight that itemset support and tag frequency are distinct concepts. When computing itemset support, we consider each sentence in a painting's description as a single transaction. This leads to a tag frequency roughly five to seven times its corresponding itemset frequency. This phenomenon leads to no high-performance models when considering the minimum support of 0.1 and without stopwords. This happens mostly because all models are either statistically similar to random guessing or lie above the 0.8 frequency threshold. We can also verify that all performant tags found without stopwords and within the minimum support of 0.05 and 0.01 are the same given the equal performance metrics. The increase of minimum support from 0.01 to 0.05 filters the leftmost cloud of data points that are inferior to the 0.2 support threshold, as shown in Figure \ref{fig:valfreq}. We obtain an optimum evaluation scenario with the 0.05 minimum support threshold mark, in accordance with previous experiments.

\begin{table}[h]
\centering
\begin{tabular}{llrrrr}
\hline
\multicolumn{2}{l}{Min. Support} & P & R & $F_1$ & $RatioF_1$ \\ \hline\\[0.05ex]
\multicolumn{1}{l}{\multirow{2}{*}{\textit{0.10 (with stopwords)}}} & All tags & .805 & .941 & .866 & 1.077 \\
\multicolumn{1}{l}{} & Performant & .758 & .892 & .818 & 1.119 \\
\multirow{2}{*}{\textit{0.05 (with stopwords)}} & All tags & .601 & .696 & .641 & 1.074 \\
& Performant & .666 & .826 & .731 & 1.154 \\ \hline\\[-0.5ex]
\multicolumn{1}{l}{\multirow{2}{*}{\textit{0.10 (no stopwords)}}} & All tags & .739 & .857 & .789 & 1.075 \\
\multicolumn{1}{l}{} & Performant & - & - & - & - \\
\multicolumn{1}{l}{\multirow{2}{*}{\textit{0.05 (no stopwords)}}} & All tags & .350 & .383 & .363 & 1.106 \\
\multicolumn{1}{l}{} & Performant & .656 & .794 & .715 & 1.409 \\
\multicolumn{1}{l}{\multirow{2}{*}{\textit{0.01 (no stopwords)}}} & All tags & .153 & .194 & .167 & 1.217 \\
\multicolumn{1}{l}{} & Performant & .656 & .794 & .715 & 1.409 \\ \hline
& & & & &
\end{tabular}
\caption{Performance comparison when considering models trained on all tags or only suitable ones.}
\label{tab:oneres}
\end{table}

\subsection{Model explanation}

\begin{figure*}[t]
\centering
\begin{subfigure}{0.4\linewidth}
\includegraphics[width=0.9\linewidth]{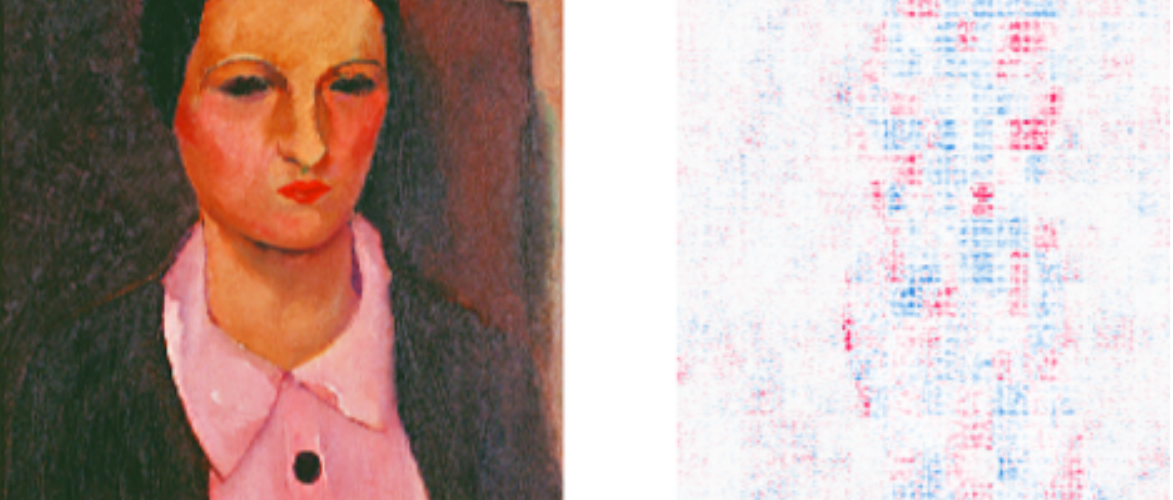}
\caption{\textit{``eyebrown,cheek,mouth''}}
\end{subfigure}
\begin{subfigure}{0.4\linewidth}
\includegraphics[width=0.9\linewidth]{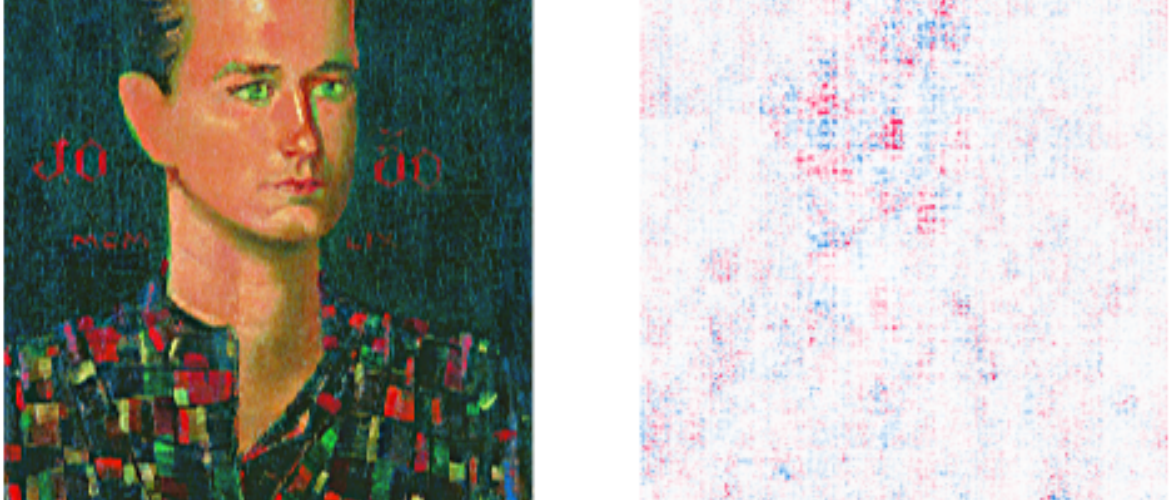}
\caption{\textit{``cheek,mouth,nose''}}
\end{subfigure}
\\[2ex]
\begin{subfigure}{0.4\linewidth}
\includegraphics[width=0.9\linewidth]{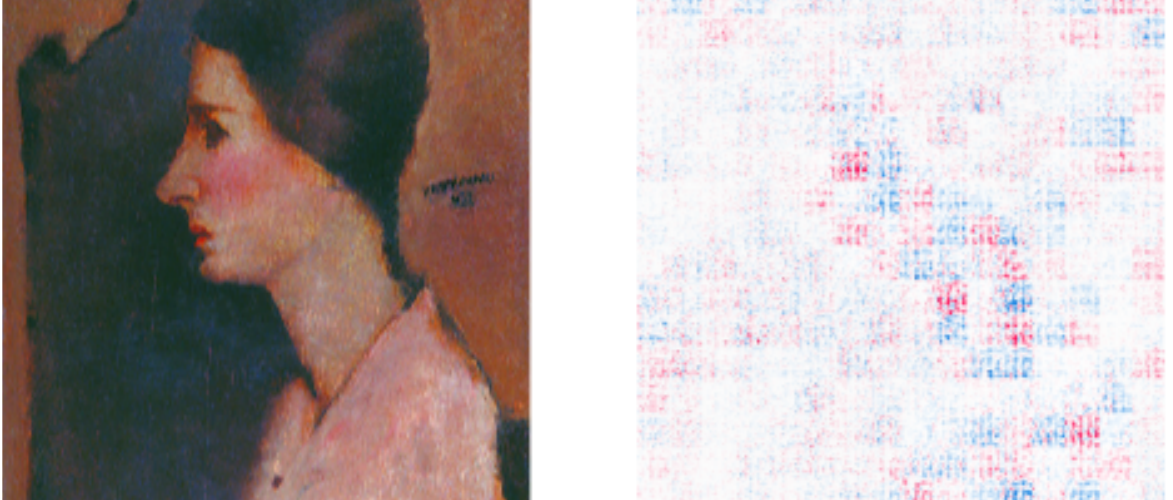}
\caption{\textit{``cheek,mouth,nose''}}
\end{subfigure}
\begin{subfigure}{0.4\linewidth}
\includegraphics[width=0.9\linewidth]{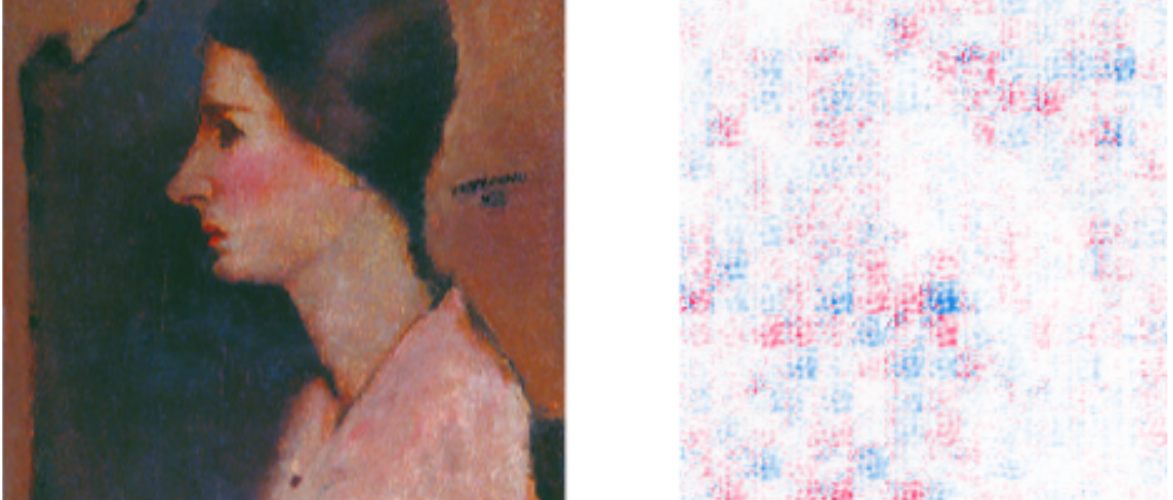}
\caption{\textit{``area,composition''}}
\end{subfigure}
\caption{SHAP gradient analysis for different paintings and tag sets.}
\label{fig:shaps}
\end{figure*}

As a final validation of the proposed method, we employ the SHAP algorithm \cite{shap2017} to trace back the gradients of the classifiers. This allows us to visualize which areas of the input images are being focused upon and driving the output prediction. Figure \ref{fig:shaps} shows the output explanation for models trained on different tags that are correlated to facial features, such as \textit{``eyebrow,cheek, mouth''}. We can observe that when an image contains a painting of a person, the models primarily focus on features found on its face.

\section{Results and discussion}

For very low values of minimum support, we have an explosion of the variance in the number of tags per painting. This happens because we have paintings with an excessive amount of specific tags while others are limited to the most frequent ones. This result suggests that using minimum support values lower than $0.01$ are not effective. Similarly, for very high support values we have no useful patterns. Values greater than or equal to $0.1$ are also not effective. We inspected the corresponding association rules as a way of ascertaining the quality of the found itemsets. For low minimum support values, we have a large presence of patterns with only one term and with lift$=1$. For high minimum support values, we have an explosion of lift.

Regarding the sizes of the found itemsets, for very high values of minimum support virtually all common sets are rather simple, containing a single element. Interestingly, more complex patterns only start appearing when there is a reduction of the minimum support to at least $ 0.05$. This is somewhat expected. Minimum relative support of $0.1$ implies that only patterns that occur in at least $10\%$ of the descriptions are considered. Due to the enormous diversity among the themes addressed in Portinari's paintings, we require significantly lower values to be able to catch diverse tags.

\begin{figure}[ht!]
\centering
\tcbox[colback=white]{\includegraphics[width=7cm]{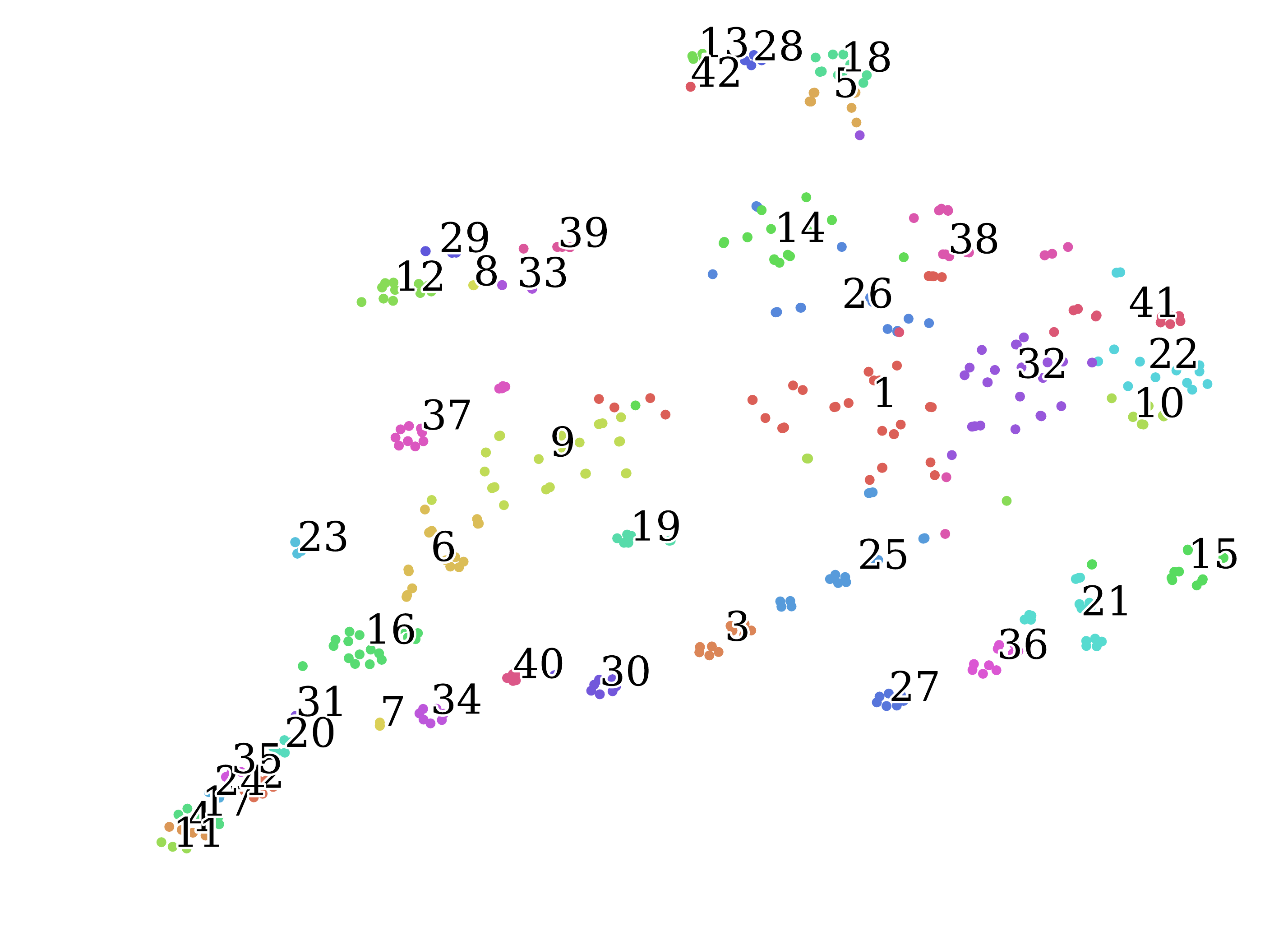}}
\caption{t-SNE reduction with 42 clusters of the Autoencoder-MLP* model and minimum support set to 0.05.}
\label{fig:tsne42}
\end{figure}

While attempting to reduce the tag-space size, we were able to successfully reduce the number of possible tags while retaining a somewhat similar performance. We observe a correlation between $F_1$ and silhouette. Large average silhouette values are associated with large $F_1$ values, while low ones yield to slightly worse classifiers. However, scenarios with near-zero silhouette values are not suitable for the proposed tag-space reduction as shown in the experiment with minimum support of $0.01$. A visual inspection of the t-SNE representation \cite{tsne2008} of the $0.05$ scenario showed in Figure \ref{fig:tsne42} reveals that there are well-defined clusters and a clear division among most of them, with little to no overlap. Upon further inspection, we verified that the clusters indeed summarize some sort of semantic or syntactic theme. For instance, cluster 19 is associated with descriptions regarding the concept hair, containing words such as ``hair'', ``straight'' and ``fringe''.

Models trained on the one-to-many scenario exploit the interdependency between itemsets and learn shared weights. If a given painting contains a frequent tag-set $\{x_1,x_2, \ldots,x_n\}$, then it must also contain all permutations of size up to $n-1$ of all the tags. This is what allows the misclassification of a superset or subset of the ground-truth to have a small negative effect on training. However, when dealing with one-to-one classifiers, each tag is trained independently and the network cannot take advantage of these dependencies. While this prevents our approach for meta-tag acquisition, it allows the analysis of each tag in a case-by-case manner, and we can strictly filter out tags that do not generalize well.

An interesting pattern arises when we plot the loss of each model versus the frequency of each tag. For high support values, we have tags that appear in virtually all paintings. Any reasonably trained model would learn to always predict these tags leading to loss close to zero. Likewise, for low support values the models should learn to seldom predict rare tags. This leads to a concave curve with a turning point near the $0.5$ frequency mark. With this in mind, we can safely reason that there exists a relationship between frequency and validation loss which is inherent to the data. A prediction according to tag frequency indicates that models seem to be learning the distribution of the tags themselves rather than meaningful features from data. This leads to models that are comparable to random guessing and unsuitable for consideration in further analysis. Therefore, we wish to focus on the models that diverge from the overall trend.

At first glance, we might be inclined in thinking that higher thresholds of minimum support lead to better classifiers, as indicated in the $F_1$ column in Table \ref{tab:oneres}. However, we should also take into account that the tag distribution varies as we consider more frequent itemsets. In the extreme case, when considering a tag that is present in all paintings, a model should achieve a $F_1$ score of $1.0$. This is true both to the experiments involving the prediction of a single tag and the whole tag vector. The $ratioF_1$ metric we proposed helps evaluating the actual utility of each model by stating how much better it is in comparison to the random baseline. In the extreme case mentioned, we would expect a $ratioF_1$ of $1.0$ meaning that the model is no better than random guessing. This metric helps us in verifying that filtering the models as proposed leads to increased performance. It also helps us in identifying an optimal minimum support threshold. When comparing the results between $0.01$ and $0.05$ frequencies, we get mostly the same performant tags. This means that further reducing the frequency would lead to no improvements. When comparing the $0.05$ and $0.10$ experiments, we observe that the increase in frequency leads to worse performant models.

\section{Conclusion}

In this work we studied the task of automatic recommendation of tags in artwork paintings by mixing frequent itemsets and deep neural networks. One of the main challenges of the classification task is obtaining ground-truth data for training the model. There are tags with similar semantics but that are associated with different paintings. We proposed two modeling approaches to solve this task: either pre-training a classifier over all tags and performing clustering on its output to obtain a group of meta-tags or training a single classifier for each candidate tag and filtering unsuitable ones.

In the first approach, we were able to achieve high-performance gains while also largely reducing the tag-space size, reaching a $F_1$ of $+.85$. For small reductions in the tag-space, we verified a drop in performance. Overall, we verified that there is a relationship between cluster silhouette and model performance, and we can keep reducing the tag-space as long as the silhouette score keeps increasing. We cannot apply the proposed cluster approach to the models trained to predict a single tag. Nevertheless, we present a different method for extracting meaningful tags by considering the loss found during training and focusing on models that diverge from the overall trend. The tags selected are highly informative, lead to high performance and contain human-understandable explanations, which are all desired characteristics.

\bibliographystyle{IEEEtran}
\bibliography{biblio}

\end{document}